\documentclass[11pt]{article}

\usepackage[]{EMNLP2023}

\usepackage{times}
\usepackage{latexsym}

\usepackage[utf8]{inputenc}

\usepackage{microtype}

\usepackage{inconsolata}

\usepackage{tabularx}
\usepackage{longtable}
\usepackage{graphicx}
\usepackage{verbatim}

\title{Blackbird language matrices (BLM), a new task for rule-like generalization  in neural networks: Motivations and Formal Specifications}

\author{Paola Merlo\\
         University of Geneva\\ Paola.Merlo@unige.ch}

\begin{document}
\maketitle

\begin{abstract}
We motivate and formally define a new task for fine-tuning rule-like generalization in large language models.

It is conjectured that shortcomings of current LLMs are due to a lack of ability to generalize.
It has been argued that, instead, humans are better at generalization because they have a tendency at extracting rules from complex data. 
We try to recreate this tendency to rule-based generalization.

When exposed to tests of analytic intelligence, for example the visual RAVEN IQ test, human problem-solvers identify the relevant objects in the picture and their relevant attributes and reason based on  rules applied to these objects and attributes. Based on the induced rules, they are able to provide  a solution to the test.

We propose a task that translates this IQ task into language.
    In this paper, we provide the formal specification for the task, 
    and the generative process of its datasets.
\end{abstract}

\section{Introduction}
Current consensus about LLM is that to reach better, possibly human-like, abilities in  abstraction and generalisation, we need  to develop tasks and data that help us understand their current generalisation abilities and help us train LLMs to more complex and compositional skills.

Generalisation in NLP has been defined in a very narrow way, as extension from a set of data points to new data points of exactly the same nature (i.i.d. assumption).
Even under this very narrow definition, recent studies show that current algorithms 
do not generalise well
\citep{belinkov-bisk2018,belinkov-glass2019tacl,kiela-etal-2021-dynabench}.

In contrast, humans are good generalizers.
  A large body of literature of experimental work has demonstrated that the human mind is predisposed to extract regularities and generate rules from data, in a way that is distinct from the patterns of activation of neural networks \citep{lakretz2019emergence,lakretz2021mechanisms,sable-meyer-ea2021}.  
  
One possible approach to develop more robust methods, then, is to pay more attention to the decomposition of complex observations and to the \textbf{ }causal chains in the generative process that gives rise to the data. 
\citet{scholkopf-etal2012}
give the example of  a dataset that consists of altitude A and average annual temperature T of weather stations
\citep{peters-ea2017} and argue that a robust model represents not their correlation, but the underlying problem structure and the causal effect of A on T. 

   To discover the underlying problem structure, machine learning research in vision has developed  the notion of \textit{disentanglement.}
A disentangled representation can be defined as one where single latent units are sensitive to changes in single generative factors, while being relatively invariant to changes in other factors \citep{bengio-ea2013}.

Let's look at an illustrative example of complex linguistic relations that might require discovering the underlying causal relations:  the \textsc{causative} alternation in English, shown in (1).

{
\begin{tabular}{llll}
(1) &The teacher & opened &  the door.    \\
    & \textsc{Agent} &            &\textsc{Theme}    \\
    & The door & opened.  & \\
       &\textsc{Theme}  &              & 
\end{tabular}}

This alternation applies to change of state verbs, such as {\it open, break, melt, burn} among many others, verbs that describe a change that affects the state of the undergoing participant (it is, after the action, \textit{the door} that changes from a state of being closed to a state of being open).
They occur in two subcategorisation frames that are related to each other in a regular way: the object of the transitive frame is the subject of the intransitive frame. This way, in terms of semantic roles, the subject of the transitive is an \textsc{agent}, but the subject of the intransitive is a \textsc{theme.}

To learn the structure of such a complex alternation automatically, a neural network must be able (i) to  identify  the elements  manipulated by the alternation, (ii) to  identify  the relevant attributes of these elements and (iii) recognize and formulate the rules of change of these attributes and elements across the two sentences. Thus finding the solution requires abilities in handling phenomena that span more than one sentence, whose correspondences must be recognised, in structure and meaning.

To  study what factors and models lead to learning  more disentangled linguistic representations, representations that reflect the underlying linguistic rules of grammar, we take the approach of developing curated data on a large scale, building models to learn these data and investigating the models' behaviour. 

To this end we develop a new linguistic task, (similar to Raven's progressive matrices for vision), which we call Blackbird Language Matrices (BLMs) that define  prediction tasks to learn these complex linguistic  patterns and paradigms. The contribution of this work lies in the definition of a new challenging learning task,  which requires  tackling 
a mixture of language processing abilities and abstract rule learning. This task takes us closer to investigations of human linguistic intelligence.

In this work, we provide the formal specification for the matrices. One such BLM problem has already been used in \citet{merlo-ea2022arxiv} and \citet{nastase-merlo2023} and its data is described in detail in 
\citet{an-etal-2023-blm}.

\subsection{Raven's Progressive Matrices for vision }

\begin{figure}
  \center
\includegraphics[width=\linewidth]{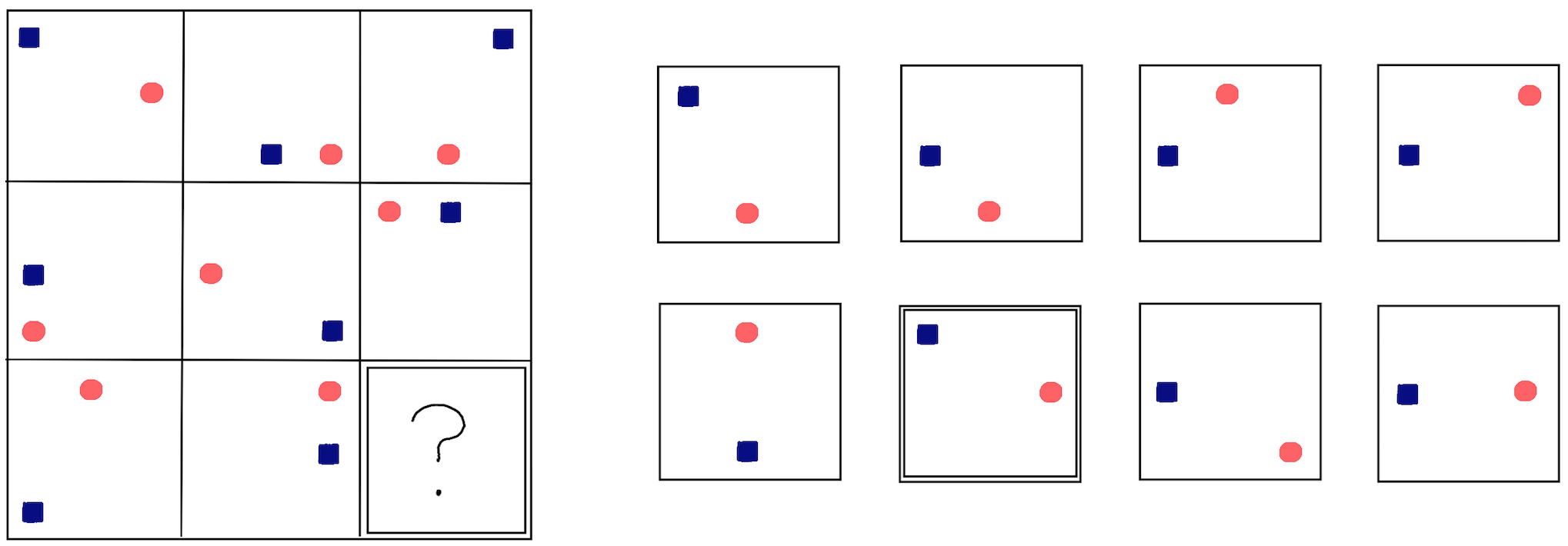}
  \hfill
  \caption{Example of progressive matrice in the visual world. The task is to determine the missing element in a visual pattern.  Given the matrix on the left, choose the last element of the matrix from the choice of elements on the right.
The matrix is constructed according to two rules (see text for explanation).
 Identifying these rules leads to the correct answer (marked by double edges).}
 \label{RPM}
\end{figure}

Raven's progressive matrices (progressive because tasks get harder) are IQ tests consisting of a sequence of images, called the \textit{context}, connected in a logical sequence by underlying generative rules \citep{raven1938}.  
The task is to determine the  missing element in this visual sequence, the \textit{answer}. The answer is chosen among a set of closely or loosely similar alternatives. An instance of this task is given in Figure~\ref{RPM}: given a matrix on the left, choose the last element of the matrix from a choice of elements. The matrices are built according to  generative rules that span the whole sequence of stimuli and the answers are constructed to be similar enough that the solution can be found only if the rules are identified correctly. For example in Figure~\ref{RPM}, the matrix is constructed according to two rules: Rule 1: from left to right, the red dot moves one place clockwise each time. This pattern continues onto the next row; Rule 2: from top to bottom, the blue square moves one place anticlockwise each time. This pattern continues onto the next column. Identifying these rules leads to the correct answer, the only cell that continues the generative rules correctly.

To establish how to develop human-like rule-based inference, it pays to survey how humans solve this kind of intelligence test and develop a corresponding formalisation.

\section{Formal Specifications of BLMs}
The formal specifications we develop in the following sections use a vocabulary that is inspired by the cognitive primitives identified  in the  study  by \citet{carpenter-ea1990}. We are also inspired by computational methods in vision,  that reproduce RAVEN's matrices, although the vision problems have slightly different properties and can arguably be described fully by context-free processes \cite{wang2015ijcai,barrett2018pmlr,zhang2019cvpr,zhu2006sig}.

We define here a new task and data format, we call it \textsc{Blackbird's Language Matrices} (BLMs).

\begin{description}
    
\item
\textsc{Definition}
Let a 4-tuple $(LP,C,W,w_c)$ be given, where 
$LP$ is the definition of the linguistic grammatical phenomenon, 
$C$ is the corresponding context matrices,
$W$ is the answer set, and $w_c$ is the selected item of $W$ that is correct.

\item
The  BLM \textit{task}  can be defined as  the instruction:

find  $(w_c \in W)$  given  $C.$
\end{description}

\noindent
A BLM \textit{problem} $(LP, C, W, Aug)$ is an instance of a BLM task, where  $Aug$ is the augmentation method for the matrices. We describe all components in the next sections.

\begin{figure*}
\begin{description}

\item[\textsc{BLM task:}] Find  $(w_c \in W)$  given  $C$,

given a 4-tuple $(LP,C,W,w_c)$, where 
$LP$ is the definition of the linguistic grammatical phenomenon, 
$C$ is the corresponding context matrices,
$W$ is the answer set, and $w_c$ is the selected item of $W$ that is correct.

\item[\textsc{BLM problem:}] A $BLM$ \textit{problem} is a tuple$(LP, C, W, Aug)$. It is an instance of a BLM task, where  $Aug$ is the augmentation method for the matrices.

\item[\textsc{BLM matrix:}]
A $BLM$ \textit{matrix} is a tuple $(S,R,T)$ s.t.
$S$ is the shape of the matrix,
$R$ are the relational operators that connect the items of the matrix,
$T$ is the set of items of the matrix.

\item[\textsc{Linguistic phenomenon LP:}] A \textit{linguistic phenomenon LP} is exhaustively defined by a grammar 
$G_{LP} = (O,A,E,I,L)$ s.t.
$O$ is the set of objects,
$A$ is the set of attributes of the objects in $O$,
$E$ is the set of external observed rules,
$I$ is the set of unobserved internal rules,
$L$ is the lexicon of objects in $O$, attributes in $A$, and operators in $E  \cup I. $

\item[\textsc{Shape}:]
$S(n,l)$ is the shape of the matrix, which consists of $n$ items and each item can be at most of length $l$.

\item[\textsc{Relational operations}:] \textsc{Alternation} $(o=NP; a_i=(s,p); i=1,2,3,...).$ 
\textsc{Progression} applies to countable attributes or ordinal attributes.

\item[\textsc{Context set:}] 
The \textit{context set $C$ }is  a set of items generated by $LP.$

\item[\textsc{Answer set:}] 
The \textit{answer set} $W$ is  a set of items generated by $LP.$ One item in W, $w_c$, is the correct answer. The other items are  the contrastive set. They are items  that violate $G_{LP}$ either in $E, I$ or in $R$. 

\item[\textsc{Augmented BLM:}] An \textit{augmented BLM} is a quadruple $(S,R,T,Aug)$. 
$S$ is the shape of the matrix, $R$ are the relational operations that connect the $T$ items of the matrix. The items $T$ are defined by $G_{LP} = (O, A, E, I, L)$ and they are drawn from the set $\mathcal{T}$. 

$Aug$ is a set of operations defined to augment the cardinality of $\mathcal{T}$, while keeping S and R constant. $Aug$ is defined by controlled manipulations of Os and As in T to collect similar elements, s.t. 

$|P(o_{aug}) - P(o)| \leq rank(10)$ and $|P(t_{aug}^i) -  P(t^i)| \leq \epsilon$ 
\end{description}
\caption{Summary of definitions concerning the Blackbird Language Matrices task.}
\label{definitions-summary}
\end{figure*}

\subsection{Defining the linguistic phenomenon}

\begin{figure}
\small
\begin{tabbing}
\textsc{Subject-verb agreement}\\
\textsc{E:  } \\
the \textcolor{red}{subject} and the \textcolor{red}{verb} \textcolor{blue}{match} in \textcolor{teal}{agreement features}.\\
\textsc{I:  } \\ 
occurs independently of distance of subject and verb.\\
\\
\textsc{Causative alternation}:\\
\textsc{E:  } \\
the \textcolor{red}{object} of alternant 1 \textcolor{blue}{becomes} \textcolor{red}{ subject} in alternant 2.\\
\textsc{I:  } \\ 
Expression of \=  thematic roles and argument structure:\\
Subject of transitive is Agent\\
Object of transitive is Theme\\
Subject of intransitive is Theme
\end{tabbing}
    \caption{Example of corresponding to formal definitions of E and I for two LPs.}
    \label{fig:FormalDefinition}
\end{figure}

The first step in the definition of the problem consists in formally defining the linguistic grammatical phenomenon as a paradigm.

\begin{description}
    
\item
\textsc{Definition}
Let a linguistic phenomenon LP be given. LP is exhaustively defined by a grammar 
$G_{LP} = (O,A,E,I,L)$ s.t.

$O$ is the set of objects

$A$ is the set of attributes of the objects in $O$

$E$ is the set of external observed rules

$I$ is the set of unobserved internal rules

$L$ is the lexicon of objects in $O$, attributes in $A$, and operators in $E  \cup I. $

\end{description}


For example, as shown in the example Figure \ref{fig:FormalDefinition}, in the subject-verb agreement phenomenon, the agreement rule is the primary production in $E$, while the fact that agreement can occur independently of the distance of the elements expresses the fact that agreement applies to structural representations, a rule in $I$.
Sometimes, but not always, I acts as a confusing factor.

Rules are triples of objects (shown in red), attributes (in green) and operations (in blue). Objects are usually phrases, attributes are usually morpho-syntactic properties of the phrases and operations are typical grammatical operations: feature match, movement (becomes), lexical substitution (changes).

\subsection{Defining the matrices}

\begin{description}
    
\item
\textsc{Definition} 
A $BLM$ \textit{matrix} is a tuple=$(S,R,T)$ s.t.

$S$ is the shape of the matrix 

$R$ are the relational operators that connect the items of the matrix

$T$ is the set of items of the matrix.

\item \textsc{Shape}

$S(n,l)$ is the shape of the matrix, which consists of $n$ items and each item can be at most of length $l$.%
%

The length of the items can vary. The items can be sentences or elements in a morphological paradigm. The choice of $n$ depends on how many items need to be shown to illustrate the paradigm and on whether the illustration is exhaustive or sampled. For example, a matrix of size eight is exhaustive for an agreement problem with three noun phrases and a two-way number differentiation (singular, plural), but can only present a sample of the information for the causative alternation.

\item \textsc{Relational operations}

Connective sequential operations, such as alternation or progression are chosen. The point of these relational operations is to trasform a list of items (sentences or words) into a predictable sequence that connects all the items. 

The values of $R$, so far, are alternations or progression. They could also be conjunction, disjunction, exclusive OR and other logical or graded operators.

\item \textsc{Alternation} applies to a given $(o,a) $ pair and loops over all the values of $a$ with a given increment defined over the items of the matrix. For example, the grammatical feature number is binary in certain languages. So, \textsc{alternation} $(o=NP; a_i=(s,p); i=1,2,3,...).$ This is used to create different alternations of $(o,a)$ in the sentence, which in the subject-verb agreement BLM is used to show independence from linear distance.

\item \textsc{Progression} applies to countable attributes or ordinal attributes, for example, existence. So, one can have 1,2,...,$n$ of a given object $n$. Progression can also apply to $position$ or to graded properties such as $length$.

\item \textsc{Items}

The items $T$ are defined by $G_{LP} = (O, A, E, I, L)$ and they are drawn from the set $\mathcal{T}$. 

\end{description}

The matrix is created by sampling $(o,a,r)$. The ways in which $r \in R$ can apply to a given $(o,a)$ pair has to be predefined, as it is not entirely context-free.

\subsection{Defining the answer set}

\begin{figure}
\begin{tabbing}
\textsc{Subject-verb number agreement}\\
Violation of E: \= wrong subject-verb  agreement\\
Violation of I: \> wrong agreement on N2 or N3\\
Violation of R: \> wrong number of attractors
\end{tabbing}
\caption{Example answer set.}
\label{fig:answers}
\end{figure}

The answer set $W$ consists of a set of items like those in $C.$ One item in W, $w_c$, is the correct answer to complete the sequence defined by $C$. The other items are  the contrastive set. They are items  that violate $G_{LP}$ either in $E, I$ or in $R$. In other words, the contrastive set comprises elements that violate one or more of  the rules of construction of the context matrix C, either in the primary rules $E$, in the auxiliary rules $I$, or in the matrix operators $R$.

Sometimes they are built almost automatically, sometimes by hand. The cardinality of the answer set is determined by how many facets of the linguistic phenomenon need to be shown to have been learned.

\subsection{Augmenting the matrices}

Different levels of lexical and structural complexity can be obtained by changing the lexical items (step by step), in a given matrix. 

\begin{description}
    
\item
\textsc{Definition} An \textit{augmented BLM} is a quadruple $(S,R,T,Aug)$. 

$S$ is the shape of the matrix, $R$ are the relational operations that connect the $T$ items of the matrix. 

\item
$Aug$ is a set of operations defined to augment the cardinality of $\mathcal{T}$, while keeping S and R constant. $Aug$ is defined by controlled manipulations of Os and As in $\mathcal{T}$ to collect similar elements, s.t. 

$|score(o_{aug}) - score(o)| \leq rank(10)$ and 

$|score(t_{aug}^i) -  score(t^i)| \leq \epsilon$ 

\end{description}

In words and in practice, we augment the sentence set $\mathcal{T}$ by modifying the noun phrases of the items in $T$. We generate alternatives with a language model choosing among the top $n$, and verifying that the acceptability score  of the resulting sentences is still acceptable, within a margin from the original sentence. The margin is set with a variable-size window and collects the top 10 alternative noun phrases. The acceptability of the resulting sentences is validated manually.

\section{BLM Formal Template and Example}
\begin{figure}[h!]
    \footnotesize
{\centering
\begin{tabular}{llll} 	\hline

\multicolumn{4}{c}{Template for contexts} \\	\hline
1 $(o_1,a_1^{v_1})$ & $(o_2,a_1^{v_1})$ & & $(o_n,a_1^{v_1})$\\
2 $(o_1,a_1^{v_2})$ & $(o_2,a_1^{v_1})$ & & $(o_n,a_1^{v_2})$\\
3 $(o_1,a_1^{v_1})$ & $(o_2,a_1^{v_2})$ & & $(o_n,a_1^{v_1})$\\
4 $(o_1,a_1^{v_2})$ & $(o_2,a_1^{v_2})$ & & $(o_n,a_1^{v_2})$\\
5 $(o_1,a_1^{v_1})$ & $(o_2,a_1^{v_1})$ & $(o_3,a_1^{v_1})$& $(o_n,a_1^{v_1})$\\
6 $(o_1,a_1^{v_2})$ & $(o_2,a_1^{v_1})$ & $(o_3,a_1^{v_1})$& $(o_n,a_1^{v_2})$\\
7 $(o_1,a_1^{v_1})$ & $(o_2,a_1^{v_2})$ & $(o_3,a_1^{v_1})$& $(o_n,a_1^{v_1})$\\
8 $(o_1,a_1^{v_2})$ & $(o_2,a_1^{v_2})$ & $(o_3,a_1^{v_1})$& $(o_n,a_1^{v_2})$\\\hline
\end{tabular}
}

\vspace{0.5cm}
{\centering
\begin{tabular}{lllll} 
\hline
\multicolumn{4}{c}{Template for  answers}\\
	\hline
\multicolumn{4}{c}{Correct}\\
$(o_1,a_1^{v_2})$ & $(o_2,a_1^{v_2})$ & $(o_3,a_1^{v_1})$& $(o_n,a_1^{v_2})$ \\
\\
\multicolumn{4}{c}{Wrong structure}\\
$(o_1,a_1^{v_2})$ & $(o_2,a_1^{v_2})$ & and $(o_3,a_1^{v_1})$& $(o_n,a_1^{v_2})$ \\
\\
\multicolumn{4}{c}{Wrong number of objects (violation of I)}\\
$(o_1,a_1^{v_1})$ & $(o_2,a_1^{v_1})$ &   $(o_n,a_1^{v_1})$ \\
\\
\multicolumn{4}{c}{Wrong values of $a$s (violation of E)}\\
$(o_1,a_1^{v_1})$ & $(o_2,a_1^{v_1})$ & $(o_3,a_1^{v_1})$& $(o_n,a_1^{v_2})$ \\
\\
\multicolumn{4}{c}{Wrong values of $a$s (violation of I)} \\
$(o_1,a_1^{v_2})$ & $(o_2,a_1^{v_1})$ & $(o_3,a_1^{v_1})$& $(o_n,a_1^{v_2})$\\
\\
\multicolumn{4}{c}{Wrong values of $a$s (violation of I)}\\
$(o_1,a_1^{v_2})$ & $(o_2,a_1^{v_2})$ & $(o_3,a_1^{v_2})$& $(o_n,a_1^{v_2})$\\
\\
\hline 
\end{tabular}
}

\vspace{0.5cm}
    
$O$ = $\{NP1, NP2, NP_, V \}$

$A$ = $\{ number\}$

$E$ = $\{(NP1_{a_{v=s}}, V_{a_{v=s}}), (NP1_{a_{v=p}}, V_{a_{v=p}}\}$

$I$ = $\{(NP1_{a_{v=s}}, V_{a_{v=s}}), (NP1_{a_{v=p}}, V_{a_{v=p}})\}$

$L$ is the lexicon of objects in $O$, attributes in $A$ = $\{s,p\}$, and operators in $E  \cup I = $\{alternation, progression\}$. $

\vspace{0.5cm}
\vspace{0.5cm}
    \caption{Examples of template and progressive matrices for the number agreement problems \cite{merlo-ea2022arxiv,an-etal-2023-blm,nastase-merlo2023}.
   }
    \label{BLM-example}
\end{figure}


\begin{figure*}
\footnotesize
\begin{tabular}{p{0.48\linewidth}p{0.48\linewidth}}
\hline
\hline\textbf{Contexts} \\\hline Example & Translation\\\hline
1  La conférence  sur l’histoire  a commencé plus tard que prévu. & \textit{The talk on history has started later than expected.}\\
2 Les responsables   du droit vont démissionner. & \textit{Those responsible for the right will resign.}\\
3    L’ exposition  avec les peintures  a rencontré un grand succès. & \textit{The show with the paintings has met with great success.}\\
4    Les menaces  de  les réformes  inquiètent les médecins. & \textit{The threats of reforms worry the doctors.}\\
5   Le trousseau  avec la clé de la cellule repose sur l’étagère.& \textit{The bunch of keys of the cell sits on the shelf.}\\
6   Les études  sur l’effet de la drogue apparaîtront bientôt. & \textit{The studies on the effect of the drug will appear soon.}\\
7    La menace  des réformes  dans l’ école inquiète les médecins. & \textit{The threat of reforms in the school worries the doctors.} \\\hline
\textbf{Answers}\\\hline
Example & Translation \\\hline
1  Les nappes  sur les tables et le banquet brillent au soleil. & \textit{The tablecloths on the table and the console shine in the sun.}\\
2 \textbf{Les copines   des propriétaires de la villa dormaient sur la plage.} & \textit{The friends of the owners of the villa were sleeping on the beach.}\\
3  Les avocats   des assassins  vont revenir. &\textit{The laywers of the murderers will come back.}\\
4  Les avocats   des assassins  du village va revenir. &\textit{The lawyers of the murderers of the village will come back.}\\
5  La visite   aux palais  de l’ artisanat approchent. & \textit{The visit of the palace of the crafts is approaching.}\\
6  Les ordinateurs avec le programme des expériences sont en panne. & \textit{The computers with the program of the experiments are broken.}
\end{tabular}
\caption{Example of lexically varied contexts for the main clause contexts.  Correct answer in bold, \cite{merlo-ea2022arxiv,an-etal-2023-blm,nastase-merlo2023}. 
}
\label{lexically-varied-contexts}
\end{figure*}

The  abstract template of context and answer set, with its objects and attributes, that corresponds to the problem of subject-verb agreement  is shown in Figure~\ref{BLM-example}.
This abstract template also corresponds to  the example of matrices shown in Figure \ref{lexically-varied-contexts}.

The sequence is generated by a rule of progression of number of attractors (one and two), a rule of subject-verb agreement that alternates every sentence between singular and plural of the head noun and a rule of number of the attractors that alternates between singular and plural with a frequency period of two. Thus, the correct answer for this example is a sentence that has three noun phrases and a plural subject and plural first attractor and singular second attractor. 

\section{Conclusions}
In this paper, we have defined precisely the main concepts underlying a new multiple-choice task, Blackbird Language Matrices (BLMs), inspired by IQ test for humans. The task is based on solving a missing-element puzzle that requires identifying the correct completion of a context sequence among several minimally contrastive alternatives. 
These matrices have already shown to define a challenging task \cite{merlo-ea2022arxiv}, and a first full-fledged dataset is available and described in \citet{an-etal-2023-blm}. This task has also been shown to be useful for detailed investigation of distributed representations \cite{nastase-merlo2023}.

Because the context sequence is developed in such a way that it is the implicit definition of a linguistic phenomenon, the matrix can become a way of teaching complex structured problems to current neural networks. It is also a powerful tool to evaluate the already existing abilities of current large language models in a challenging task of analytic intelligence. Work is underway on all these aspects.

\section*{Ethics Statement}
To the best of our knowledge, there are no ethics concerns with this paper. 

\section*{Acknowledgments}

We gratefully acknowledge the partial support of this work by the Swiss National Science Foundation, through grants  \#51NF40\_180888 (NCCR Evolving Language) and SNF Advanced grant  TMAG-1\_209426 to PM. Thanks to all collaborators and audiences that have shown interest in this concept.
\newpage
\bibliography{big-frontiers,BLM21,raven_dataset}
\bibliographystyle{acl_natbib}

\end{document}